\documentclass[10pt,twocolumn,letterpaper]{article}

\usepackage{iccv}
\usepackage{times}
\usepackage{epsfig}
\usepackage{graphicx}
\usepackage{amsmath}
\usepackage{amssymb}
\usepackage{soul}
\usepackage{color}
\usepackage{booktabs}
\usepackage{multirow}
\usepackage[cal=boondoxo]{mathalfa}
\usepackage{paralist}
\usepackage{fontawesome}
\usepackage[accsupp]{axessibility}
\DeclareMathAlphabet{\pazocal}{OMS}{zplm}{m}{n}
\usepackage[pagebackref=true,breaklinks=true,letterpaper=true,colorlinks,bookmarks=false]{hyperref}

 \iccvfinalcopy 

\def\iccvPaperID{9446} 

\ificcvfinal\fi

\begin{document}



 \title{Cross-modal Orthogonal High-rank Augmentation \\ for RGB-Event Transformer-trackers}
\author{Zhiyu Zhu, Junhui Hou\thanks{Corresponding author: Junhui Hou. This work was supported in part by the Hong Kong Research Grants Council under Grant 11218121 and Grant 11202320, and in part by the Hong Kong Innovation and Technology Fund under Grant 
MHP/117/21. }, and Dapeng Oliver Wu\\
Department of Computer Science, City University of Hong Kong\\
{\tt\small zhiyuzhu2-c@my.cityu.edu.hk; jh.hou@cityu.edu.hk; dapengwu@cityu.edu.hk}
}

\maketitle

\ificcvfinal\thispagestyle{empty}\fi

\begin{abstract}
This paper addresses the problem of cross-modal object tracking from RGB videos and event data.  Rather than constructing a complex cross-modal 
fusion network, we explore the great potential of a pre-trained vision Transformer (ViT). Particularly, we delicately investigate plug-and-play training augmentations that encourage the ViT to bridge the vast distribution gap between the two modalities, enabling comprehensive cross-modal information interaction and thus enhancing its ability.
Specifically, we propose a mask modeling strategy that randomly masks a specific modality of some tokens to enforce the interaction between tokens from different modalities interacting proactively. 
To mitigate network oscillations resulting from the masking strategy and further amplify its positive effect, we then theoretically propose an orthogonal high-rank loss to regularize the attention matrix. 
Extensive experiments demonstrate that our plug-and-play training augmentation techniques can significantly boost state-of-the-art one-stream and two-stream trackers to a large extent in terms of both tracking precision and success rate. Our new perspective and findings will potentially bring insights to the field of leveraging powerful pre-trained ViTs to model cross-modal data. The code is publicly available at
\url{https://github.com/ZHU-Zhiyu/High-Rank_RGB-Event_Tracker}.
\end{abstract}

\section{Introduction}

Event cameras asynchronously capture pixel intensity fluctuations with an ultra-high temporal resolution, low latency, and high dynamic range, making it gain increasing attention recently~\cite{mitrokhin2018event,perot2020learning,gallego2020event}. Owing to such admirable advantages, event cameras have been widely adopted in various applications, such as object detection~\cite{mitrokhin2018event, li2022asynchronous, mondal2021moving, perot2020learning, de2020large} and depth/optical flow estimation~\cite{gallego2018unifying,zhu2019unsupervised}. Particularly, the distinctive sensing mechanism  makes event cameras to be a promising choice for object tracking~\cite{ramesh2020tld,kueng2016low,zhang2021object,zhulearning,gehrig2018asynchronous,gehrig2020eklt}. 

 Despite many advantages of event-based object tracking under special environments, e.g., low-light, high-speed motion, and over-exposed,  event data lack 
 sufficient visual cues, such as color, texture, and complete contextual appearance that can be easily captured by RGB data, 
 resulting in only event-based vision still suffering from relatively inferior performance in practice. Thus, a more promising direction is to investigate cross-modal object tracking  from both RGB and event data, where the merits of  the two modalities can be well leveraged for pursuing higher performance. 
However, the vast distribution gap between RGB and event data poses significant challenges in designing algorithms for modeling cross-modal information. 
Most existing pioneering cross-modal trackers heavily engage in robust cross-modal fusion modules, which is cumbersome to use advanced embedding backbones for boosting performance. 

In view of the success of Transformer-based tracking algorithms \cite{lin2021swintrack,zhang2022spiking,wang2021transformer,chen2021transformer,zhou2022pttr}, where the multi-head attention naturally models the indispensable correlation relationship between template and search regions, we plan to investigate the potential of pre-trained powerful vision Transformers (ViTs) in cross-modal object tracking from both RGB and event data.
However, those pre-trained Transformers with RGB data may not be able to fully model the \textit{essential} feature interaction across RGB and event data, due to the distribution gap between the two modalities.
To this end, we study \textit{plug-and-play} training techniques for augmenting the pre-trained Transformer used as the embedding backbone of our RGB-event object tracking framework.

To be specific, to promote the learning of the attention layer across two modalities, we propose a cross-modal mask modeling strategy, which randomly masks/pops out the multi-modal tokens. We anticipate that, in reaction to the absence of a particular modality at certain locations, the network would proactively enhance interactions on the remaining cross-modal tokens. Nevertheless, randomly masking tokens will inevitably alter data distributions and introduce disruptions, impeding network training. To mitigate the induced negative effect,  we further propose a  regularization term to guide the training of each attention layer. Based on the observation that the values of internal attention matrices of a Transformer indicate the degree of cross-modal feature interaction, 
we propose to orthogonalize the attention matrix to promote its rank obligatorily. Beyond, we anticipate that such regularization could encourage the cross-modal correlation to be evenly and concisely established using the multi-domain signatures, rather than unduly reliant on a specific domain. Finally, we apply the proposed techniques to state-of-the-art one-stream and two-stream Transformer-based tracking frameworks and experimentally demonstrate that their tracking performance is further boosted significantly. 




In summary, the contributions of this paper are:
\begin{compactitem}
    \item a mask modeling strategy for encouraging the interaction between the cross-modal tokens in a \textit{proactive} manner;
    \item \textit{theoretical} orthogonal high-rank regularization 
    for suppressing network fluctuations induced by cross-modal masking while amplifying its positive effect; 
    and
    \item  new state-of-the-art baselines for RGB-event object tracking.
    \end{compactitem}

Last but not least, our novel perspectives will potentially bring insights to the field of leveraging pre-trained powerful ViTs to process and analyze cross-modal data.

\vspace{0.2cm}
\section{Related Work}
\subsection{Object Tracking}
Recent years have seen remarkable progress in the study of object tracking, which is primarily due to the widespread success of deep learning \cite{bertinetto2016fully,li2019siamrpn++}. Based on the distribution of computational burdens, current methods could be generally divided into two-stream \cite{bertinetto2016fully,bhat2019learning,lin2021swintrack} and one-stream methods \cite{ye2022joint,chen2022backbone}. As the earlier invented and relatively mature ones, most offline Siamese-based tracking methods~\cite{bertinetto2016fully, li2018high, li2019siamrpn++} fall into the first category. It utilizes a delicate embedding backbone to extract semantic-rich embeddings and then models the target location via either a direct proposal head \cite{bertinetto2016fully} or an online optimization process \cite{bhat2019learning}, which is also called deep Siamese-trackers or discriminative correlation filters, respectively \cite{javed2022visual}. SiamFC~\cite{bertinetto2016fully} first developed a fully-convolutional architecture to fuse template and search embeddings for object tracking. Though introducing a single-stage RPN~\cite{ren2016faster} detector SiamRPN~\cite{li2018high} achieved target object tracking by comparing the current-frame features to those from a template. To remove the disturbance factors, e.g., padding,  SiamRPN++~\cite{li2019siamrpn++} introduced a spatial-aware sampling strategy and further utilized ResNet~\cite{he2016deep} to embed representative features for Siamese-based tracking. 
DiMP~\cite{bhat2019learning} proposed to exploit both target and background appearances to achieve object tracking. KYS \cite{bhat2020know} represented the scene information as dense state vectors and utilizes such state vectors to maximize the tracking performance. Besides, some spatio-temporal-based methods also exploit temporal information to achieve robust and effective tracking \cite{nam2016learning,jung2018real,wang2020tracking,yan2021learning}. MDNet~\cite{nam2016learning} separated domain-independent from domain-specific information via a CNN-based framework. RT-MDNet~\cite{jung2018real} further improved it via an RoI-Align strategy, which extracts more precise embeddings from feature maps of targets and candidates. Swin-tracker \cite{lin2021swintrack} introduced the Swin-Transformer~\cite{liu2021swin} to effectively encode the semantic information from input images for high-performance visual tracking.

Due to the extraordinary correlation modeling ability of Transformer, an emerging branch of one-stream methods shows strong potential in correlation modeling. OS-track~\cite{ye2022joint} unified the embedding and relation modeling  processes with a single vanilla ViT~\cite{dosovitskiyimage}, which achieves admiring performance with reduced computational resources. Meanwhile, SimViT-Track~\cite{chen2022backbone} proposed a similar approach, which feeds search and template image tokens straight into a ViT backbone and performs regression and classification on the resulting tokens.

In summary, with the success of existing embedding backbones, such as ViT~\cite{dosovitskiyimage} and Swin-Transformer~\cite{liu2021swin}, more intriguing  and effective methods have been proposed recently. While these methods could achieve admirable performance, most of them are driven by matching semantically identical segments of the search and template regions viewed as RGB images. As a result, their performance is inextricably tied to imaging characteristics, which can be  compromised in specific scenarios such as high-speed  and low-light scenes. Hence, it is highly desired to incorporate  multi-modal inputs to remedy each deficiency.  Moreover, the crucially multi-modal data necessitates additional efforts to generalize these methods to the event-based.

\subsection{Event-based Tracking}
Owing to its innate characteristics and superiority for object tracking, event-based tracking has been a progressively prevalent subject for research in recent years. Additionally, existing approaches may be broadly classified into two categories: model-based and data-driven. Through describing surrounding environments by a photometric 3D map, Bryner \textit{et al.}~\cite{Bryner2019event} proposed to track the 6-DOF pose of a camera. To capture the spatio-temporal geometry of event data, Mitrokhin \textit{et al.}~\cite{mitrokhin2018event} utilized a parametric model to compensate camera motion.  Based on a pipeline of tracking-learning-detection, Ramesh \textit{et al}.~\cite{Ramesh2018LongtermOT}  proposed an object tracking algorithm for event cameras, which is the first learning-based long-term event tracker. Then, Li \textit{et al.}~\cite{li2019robust} introduced the VGG-Net-16 to encode the appearance of the event-stream object. Inspired by the classic Siamese-matching paradigm, Chae \textit{et al.}~\cite{chae2021siamevent} presented to track objects via learning an edge-aware similarity in the event domain. Recently, Zhang \textit{et al.}~\cite{zhulearning}, introduced a spiking transformer for encoding spatio-temporal information of object tracking. Moreover, ZHU \textit{et al.}~\cite{zhulearning} proposed to utilize inherent motion information of event data to achieve effective object tracking. To summarize, although there are some promising studies that provide directive insights for event-based tracking, a limited number of works have sought to find complementary information from RGB data, e.g., semantic information.

\subsection{Cross-modal Learning}
Fusing embedding with multiple modalities is a sensible solution for perceiving and recognizing the objects robustly and accurately \cite{ramachandram2017deep, zhang2023bidirectional, zhu2021semantic}. However, for current machine learning algorithms, learning representative patterns from multiple modalities is still a challenging issue \cite{liu2022cmx,jaritz2020xmuda}. Wang \textit{et al.}~\cite{wang2021pointaugmenting} proposed to apply data augmentation techniques to boost cross-modal 3D object detection.  Liu \textit{et al.}~\cite{liu2022cmx} utilized cross-modal feature rectification and fusion models for image segmentation with input from multiple modalities. Jaritz \textit{et al.}~\cite{jaritz2020xmuda} solved the multi-modal segmentation issue from the perspective of unsupervised domain adaptation. Moreover, Wang \textit{et al.}~\cite{wang2020cross} designed an RGB-T tracking framework by propagating the intermodal pattern and long-term context. Ye \textit{et al.}~\cite{ye2019cross} proposed a cross-modal self-attention module to achieve natural language-based image segmentation via adaptively capturing informative words and important regions in images. Zeng \textit{et al.}~\cite{wang2020cross} proposed to project the camera features onto the point set on LiDAR. In summary, recent works are clearly founded on network architecture, as is evident by their prevalence. Moreover, the current advanced Transformer paradigm could adaptively process different modalities. However, there is still a lack of further investigations and analysis of the internal mechanism. 

\vspace{0.2cm}
\section{Proposed Method}
\subsection{Motivation}
Learning the correlation between the template and search regions robustly and precisely is one of the most essential aspects of object tracking. Fortunately, with current advancements in the multi-head attention mechanism, such correlation 
could be naturally achieved via Transformer-based frameworks \cite{wang2021transformer,chen2021transformer}. 
However, current powerful ViTs were usually pre-trained with RGB data, e.g., ImageNet~\cite{deng2009imagenet}, 
potentially resulting in that they cannot be adequately adapted to  cross-modal learning, i.e., the full feature interaction between RGB and event data cannot be well achieved, which is essential for cross-modal object tracking, due to the vast distribution gap between RGB and event data. Accordingly, the tracking performance may be limited. 

Instead of following existing cross-modal research paradigms mainly focused on designing sophisticated cross-modal information fusion networks, we aim to explore \textit{plug-and-play training augmentation techniques} to mitigate the above-mentioned potential limitation of a pre-trained ViT used as the embedding backbone of an RGB-Event object tracking scheme. 
Generally, based on a fundamental and essential premise that different modalities possess their own unique benefits for a cross-modal tracker, token embedding information should be adequately transmitted across multi-modalities, especially for the regions with target objects, in a bid to enhance themselves using specific merits from the other modality. Thus, we propose a mask modeling strategy to enable the network to proactively exploit the cross-modal information in Sec.~\ref{Sec:CMLearning}. Furthermore, we propose a high-rank orthogonalization mechanism in Sec.~\ref{Sec:HRPromoting}, which can not only alleviate network fluctuations induced by the mask modeling strategy but also further boost cross-modal information interaction.  

\begin{figure}
  \centering
    \includegraphics[clip, width=0.45\textwidth]{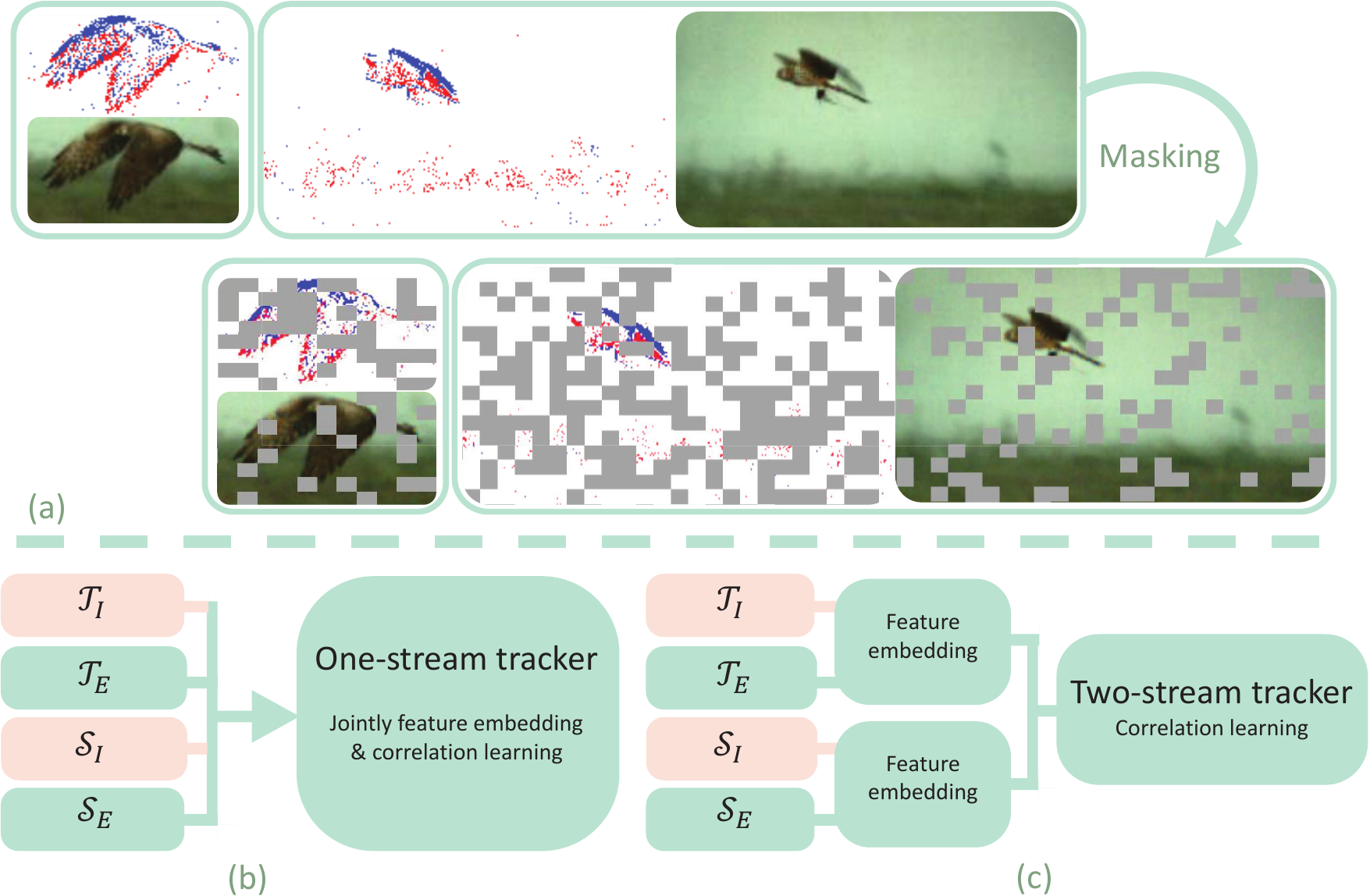}
  \caption{Illustration of (\textbf{a}) the proposed cross-modal mask modeling strategy on template and search data. General structures of Transformer-based RGB-event trackers (\textbf{b}) one-stream and (\textbf{c}) two-stream, where $\pazocal{T}$ and $\pazocal{S}$ represent the tokens of template and search patches, with subscripts $I$ and $E$ indicating the RGB and event modalities, respectively. 
  }
    \label{fig:pipeline}
\end{figure}

In what follows, we will detail the proposed techniques adapted to both one-stream and two-stream trackers, as illustrated in Fig.~\ref{fig:pipeline} (\textcolor{red}{b}) and Fig.~\ref{fig:pipeline} (\textcolor{red}{c}), respectively. 
We always use $I$ and $E$ in the subscripts to indicate the RGB and event modalities, and $\pazocal{T}$ and $\pazocal{S}$ are the tokens of template and search regions, respectively. 

\subsection{Mask-driven Cross-modal Interaction} 
\label{Sec:CMLearning}
Grouping tokens via similarity is one of the most representative steps for the self-attention mechanism of a Transformer~\cite{vaswani2017attention}. However, due to the distribution gap between 
tokens corresponding to different modalities, the similarity-driven attention may tend to aggregate information from the identical modality, hence impeding the cross-modal learning, 
Thus, how to effectively and efficiently promote the cross-modal interactions is \textit{critical}  for maximizing the potential of a pre-trained ViT for RGB-event  object tracking.

\begin{figure}
  \centering
    \includegraphics[trim={0cm 0 0 0},clip, width=0.4\textwidth]{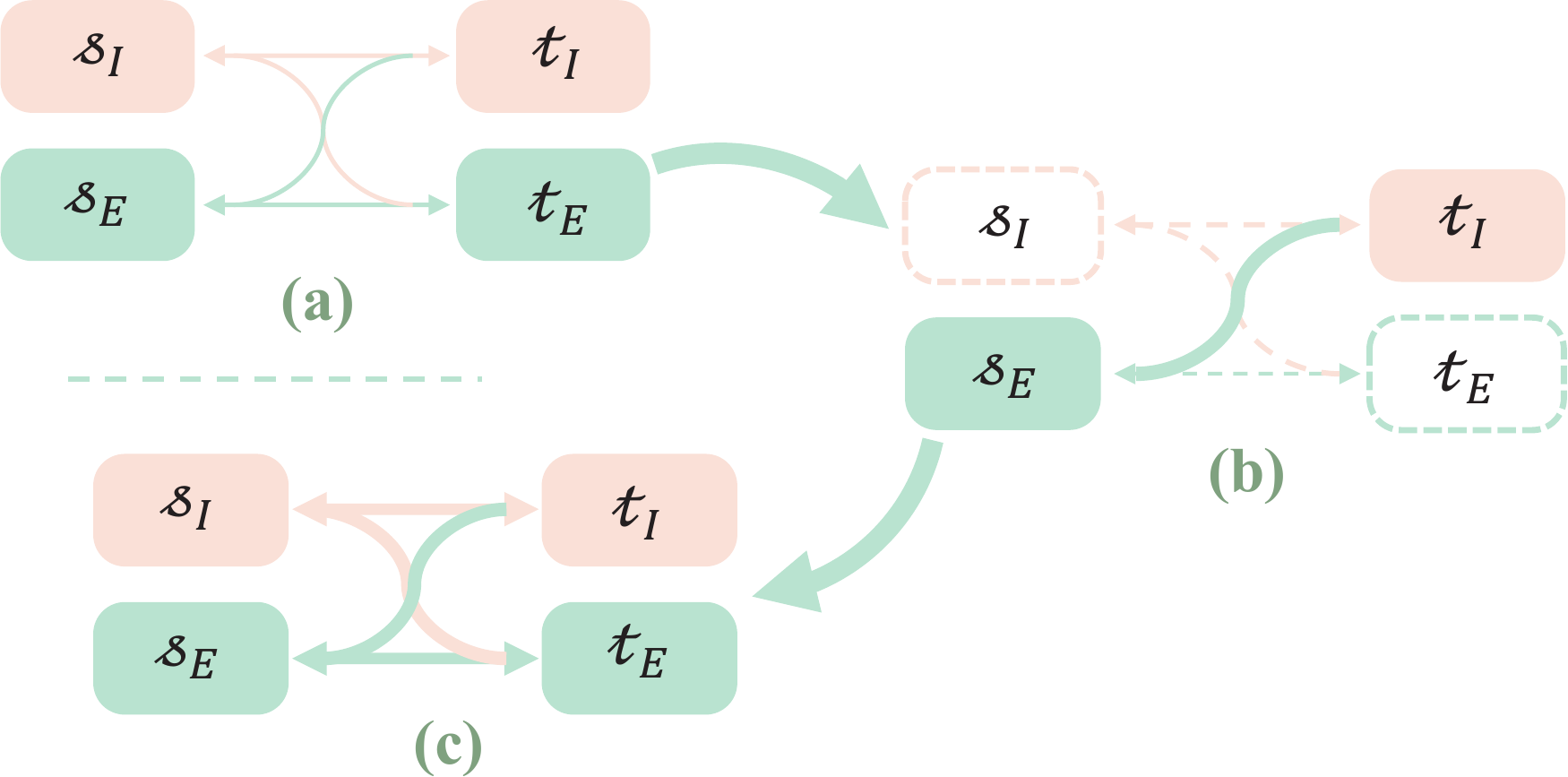}
  \caption{Illustration of the underlying intuition of our cross-modal strategy, where $\mathcal{s}$, and $\mathcal{t}$ denote tokens from search and template regions, respectively. The width of the lines indicates the degree of cross-modal interaction, i.e., the thicker, the more comprehensive. (\textbf{a}) The correlation between tokens is insignificant in the baseline model. (\textbf{b}) During training, tokens are \textit{randomly} masked to facilitate cross-modal interaction. The random manner ensures the potential token interaction routes are strengthened with an equal probability.  (\textbf{c}) The augmented model can cross the gap between cross-modal tokens to interact with each other. }
    \label{fig:mask_modeling}
\end{figure}

We propose a cross-modal mask modeling strategy to address this issue in a \textit{proactive} manner, shown as Fig.~\ref{fig:pipeline} (\textcolor{red}{a}). 
As illustrated in Fig.~\ref{fig:mask_modeling}, the underlying intuition of this strategy 
is through  removing the patches of different modalities and locations, we expect that the task loss would enforce the network to spontaneously enhance/build cross-modal correlation, due to the remaining tokens in different modalities. Once the interaction is established, the RGB and event tokens may learn to shrink the distribution gap, maintaining such correlation to the inference phase.
Specifically, we apply random masks to RGB and event data to remove distinct patches. 
To begin, for the one-stream methods, masking elements can be readily accomplished by simply popping out corresponding elements, which could concurrently lessen the  network training burden. 
For the two-stream methods, due to the large computational resource consumption of the embedding backbone, we directly average  the masked features of RGB and event data at the primary stage, which are  
further fed into the high-level embedding backbone and relation modeling modules for the object proposal.

\noindent\textbf{\textit{Remark}}. It is worth noting that the motivation and objective of the proposed masking strategy are considerably \textbf{different} from those of the well-known masked image modeling \cite{he2022masked,xie2022simmim,klenk2022masked}. We start from the pursuit of promoting the network to actively utilize cross-modal information. Thus, the patches with distinct positions across RGB and event modalities are randomly removed to permit each location can be perceived by the network but with different modalities. However, mask image modeling pre-trains network weights to comprehend image semantics by feeding just a subset of image patches to reconstruct the unseen area.

Although such a masking strategy used in the training phase is expected to strengthen the ability of the network to perceive cross-modal information to some extent, the randomly dropped information would potentially result in an unstable  training process. Moreover, such disruptions are especially devastating for one-stream algorithms, which must concurrently learn representative embeddings and establish the relationship between the cross-modal template and search tokens (see the experimental demonstration in Sec.~\ref{sec:ablation}). Thus, 
 to pull the network out of this predicament,  we further propose  orthogonal high-rank regularization in a \textit{theoretical} manner in the next section.

\begin{figure}
\definecolor{right}{RGB}{255,162,180}
\definecolor{left}{RGB}{106,182,164}
  \centering
    \includegraphics[trim={0.1cm 0 0 0},clip, width=0.48\textwidth]{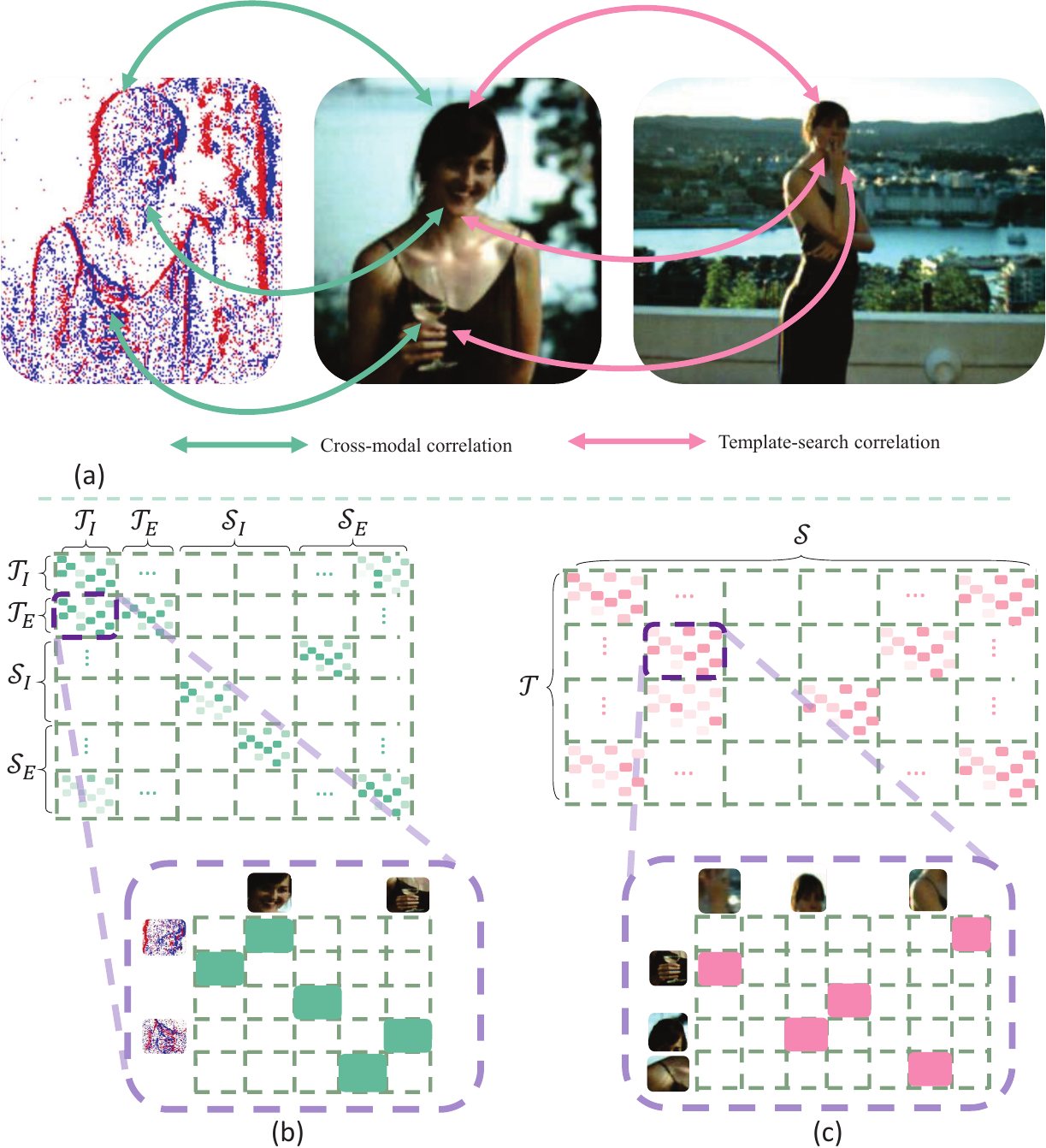}
  \caption{(\textbf{a}) Illustration of the correlation modeling for the cross-modal tracker, where \textcolor{left}{left arrows} reflect the desirable \textit{cross-modal correlation}  and \textcolor{right}{right ones} indicate the satisfactory \textit{search-template correlation}. The matrix form of such correlations is shown (\textbf{b}) and (\textbf{c}). Each row consists of attention values of a \textbf{query} to multiple \textbf{keys}, i.e., the summation of each row is equal to 1.  Here, we use the indicators of the row and column to represent the corresponding block, e.g., $M_{\pazocal{T}_E,\pazocal{T}_I}$ indicates the matrix block located at the 2$^{nd}$ row and the 1$^{st}$ column.  We also show the real matrices in Fig.~\ref{fig:attn_matrix}.}
    \label{fig:desirable}
\end{figure}
\subsection{Orthogonal High-rank Regularization} 
\label{Sec:HRPromoting}
 To appreciate the multi-head attention mechanism, we take a one-stream tracker \cite{tang2022revisiting} with the vanilla ViT \cite{dosovitskiyimage} as an example.  As illustrated in Fig.~\ref{fig:desirable} (\textcolor{red}{b}), its internal self-attention layers concurrently perceive the RGB and event tokens from both the template and search areas. Depending on the query and key belongings $k \in \mathbb{R}$, we can partition the resulting attention matrix into $k^2$ blocks (Here $k=4$). Note that the attention values of a typical block reflect the degree of the interaction between tokens.

To 
mitigate network disturbs induced by the cross-modal mask modeling strategy 
and further amplify its positive effect (i.e., boosting cross-modal learning), we concentrate on the cross-modal zones of the attention matrix, such as $M_{\pazocal{S}_I,\pazocal{S}_E}$, and $M_{\pazocal{S}_E,\pazocal{S}_I}$. Assuming that if tokens are well-embedded and with highly discriminative features, each token will form a unique correlation with its identical counterpart, resulting in each row or column being orthogonal to the others. Moreover, as attention elements are non-negative, the corresponding matrix should be  \textit{full rank}\footnote{We refer readers to the \textit{Supplementary Material} for more details}.  Therefore,  we propose the following regularization to encourage some desired blocks of the attention matrix to be high-rank: 
\begin{equation}
\label{eq:1}
    L(M,\tau) = \|\texttt{diag}(\Sigma)-\texttt{vec}(\tau)\|_1, M = U \Sigma V,
    \vspace{0.1cm}
\end{equation}
where $\tau \in \mathbb{R}$ is a pre-defined threshold value, $U\in\mathbb{R}^{n\times n}$, $\Sigma \in \mathbb{R}^{n\times m}$, and $V\in\mathbb{R}^{m\times m}$ are the outputs of the singular value decomposition (SVD) of block $M\in \mathbb{R}^{n\times m}$, and $\texttt{diag}(\cdot)$ returns a vector, consisting of the main diagonal elements of the input matrix, and $\texttt{vec}(\cdot)$ converts an input scalar to be a vector by duplicating the scalar. We impose the regularization term onto a set of blocks of the attention matrix  $\{M^{(i )}\}_{i=1}^{N}$ standing for the interaction of cross-modal tokens.  
Due to its strong regularization effect, we empirically select the blocks corresponding to image-to-event attention (i.e.$M_{\pazocal{S}_I,\pazocal{T}_E},$ and $M_{\pazocal{S}_I,\pazocal{S}_E}$), and  the blocks to event-to-image attention (i.e., $M_{\pazocal{S}_E, \pazocal{T}_I}$, and $  M_{\pazocal{S}_E, \pazocal{S}_I}$).
Moreover, as computing the SVD of a matrix is time-consuming, we randomly choose a layer to implement this regularization at each optimization step, instead of operating it in each layer.

For the two-stream methods, since the input data from different modalities are mixed in a preceding embedding backbone as shown in Fig.~\ref{fig:pipeline} (\textcolor{red}{c}), e.g., swin-Transformer \cite{liu2021swin}, the resulting attention matrix only consists of two parts, i.e., the search-to-template and template-to-search regions, as illustrated in Fig.~\ref{fig:desirable} (\textcolor{red}{c}).
Under this scenario, we anticipate that the discriminative cross-modal tokens will be able to form a unique correlation with the identical object parts across template and search areas. As shown in the right part of Fig.~\ref{fig:desirable} (\textcolor{red}{a})  and Fig.~\ref{fig:desirable} (\textcolor{red}{c}), such a relationship would also produce that each row is orthogonal to the others. 
Thus, we also regularize the regions belonging to the target objects in $M_{\pazocal{S},\pazocal{T}}$. Specifically, guided by bounding box information, we first mask the attention weights in non-target regions of $M_{\pazocal{S},\pazocal{T}}$, then apply Eq.~\eqref{eq:1} to increase the rank of the masked matrix.

\subsection{Training}
To train a Transformer-based tracker with  the proposed plug-and-play augmentation techniques, at each optimization step, we first randomly mask/pop out event and image patches with a ratio of $\delta_e$ and $\delta_i$ ($0<\delta<1$), respectively. Then, we train the whole network with the following loss function:
\begin{equation}
\label{eq:Loss}
    L_{all} = L_{task} + \alpha L(M,\tau),
    \vspace{0.1cm}
\end{equation}
where $L_{task}$ denotes the original task loss function, composed of regression and classification branches, and $\alpha$ is a balanced weight for the proposed regularization term. 




\vspace{0.2cm}
\section{Experiment}

\noindent \textbf{Implementation details}. We evaluated the proposed plug-and-play training augmentation techniques on both one-stream and two-stream trackers. We set template and search sizes as $128$ and $256$, respectively, which contain $2\times$ and $4\times$ regions than annotations. Moreover, the location and scale jitter factors of the search region are set as 3 and 0.25, respectively (No jitter to template region). For one-stream, we directly adopted the SOTA method named color-event unified tracking (CEUTrack) \cite{tang2022revisiting} as our baseline model (ViT-B). Moreover, we also adapted ViT-L to CEUTrack to further validate the effectiveness of the proposed regularization term. During training, we used the same optimizer (ADAW), learning rate scheduler, and task loss function as the original paper. We set the batch size as 24 and the augmentation weight $\alpha$ in Eq.~\eqref{eq:Loss} empirically as $1.2$. The masking ratios of both modalities $\delta_i$ and $\delta_e$ were set to $0.1$. 

For two-stream, to the best of our knowledge, there is no Transformer-based RGB-event tracker available, 
we chose the most recent event cloud-based motion-aware tracker (MonTrack) \cite{zhulearning} and modified it with the proposal head of a Transformer-tracker~\cite{chen2021transformer} and the backbone of pre-trained swin-v2~\cite{liu2022swin} to construct two-stream RGB-event trackers for the detailed architecture). Moreover, we  tested lightweight and heavy backbones, i.e., Swin-V2-Tiny~\cite{liu2022swin} and Swin-V2-Base~\cite{liu2022swin}, to achieve comprehensive evaluation, and the resulting baselines are named  MonTrack-T and MonTrack-B, respectively. To train the whole framework, we utilized the AdamW optimizer~\cite{loshchilovdecoupled} with the learning rate of $1e^{-4}$for the proposal head and $1e^{-5}$ for the backbone. We set the weight decay as $1e^{-4}$. MonTrack-T and MonTrack-B were trained with 57K and 81K steps, respectively. We empirically set the value of $\alpha$ as $1.0$, and the masking ratios of RGB and event data $\delta_i$ and $\delta_e$as $0.4$ and $0.3$, respectively.

We refer readers to the \textit{Supplementary Material} for the detailed network architectures and settings.  \\

\begin{table}
\setlength\tabcolsep{3pt}
  \caption{Quantitative comparison on the \textbf{FE108} dataset in terms of four metrics, i.e., representative success rate (RSR), representative precision rate (RPR), and overlap precision (OP) with the threshold equal to 0.5 (OP$_{0.50}$) and 0.75 (OP$_{0.75}$). For all metrics, the \textbf{larger}, the \textbf{better}. ``RawE" stands for raw event data, and ``EI" for the event image representation of event data. 
  } 
  \vspace{0.2cm}
  \label{table:FE108}
  \centering
  \resizebox{.48\textwidth}{!}{
  \begin{tabular}{l | c |cccc}
    \toprule[1.2pt]
    Methods  & Modality & RSR & OP$_{0.50}$ & OP$_{0.75}$ & RPR \\
    \midrule
    \midrule
    CLNet~\cite{dong2020clnet} & RGB & 34.4 & 39.1& 11.8& 55.5\\
    KYS~\cite{bhat2020know} & RGB & 26.6&30.6&9.2&41.0\\
    ATOM~\cite{danelljan2019atom} & RGB & 46.5 & 56.4 &20.1 &71.3\\
    PrDiMP~\cite{danelljan2020probabilistic} & RGB & 53.0 & 65.0& 23.3& 80.5\\
    
    FENet~\cite{zhang2021object} & EI & 53.2 & 61.4& 19.8& 80.0\\
    MonTrack~\cite{zhulearning} & RawE & 54.9 & 65.8& 21.4& 85.9\\
    
    ATOM~\cite{danelljan2019atom} & RGB + EI & 55.5 & 70.0& 27.4 & 81.8\\
    DiMP~\cite{bhat2019learning}  & RGB + EI & 57.1 &71.2 & 28.6 & 85.1\\
    PrDiMP~\cite{danelljan2020probabilistic} & RGB + EI & 59.0 & 74.4& 29.8& 87.7\\
    FENet~\cite{zhang2021object} & RGB + EI & 63.4 & 81.3& 34.3& 92.4\\
    \midrule
    MonTrack-T & RGB + RawE & 63.3 &82.9 &37.2&90.7 \\
    MonTrack-T+Ours & RGB + RawE & 66.3 & 86.4 & 40.0 &95.3\\
     \textbf{Improvement}       & -- -- &\textcolor{red}{+3.0} &\textcolor{red}{+3.5} &\textcolor{red}{+2.8} &\textcolor{red}{+4.6}\\
    \midrule
    MonTrack-B & RGB + RawE & 64.3 & 84.7 & 35.3 &93.2\\
    MonTrack-B+Ours & RGB + RawE & 68.5 & 89.4  &45.4 &96.2  \\
    \textbf{Improvement}      & -- -- &\textcolor{red}{+4.2} &\textcolor{red}{+4.7} &\textcolor{red}{+10.1} &\textcolor{red}{+3.0}\\
    \bottomrule[1.2pt]
  \end{tabular}
  }
\end{table}
\begin{figure}
  \centering
    \includegraphics[width=0.5\textwidth]{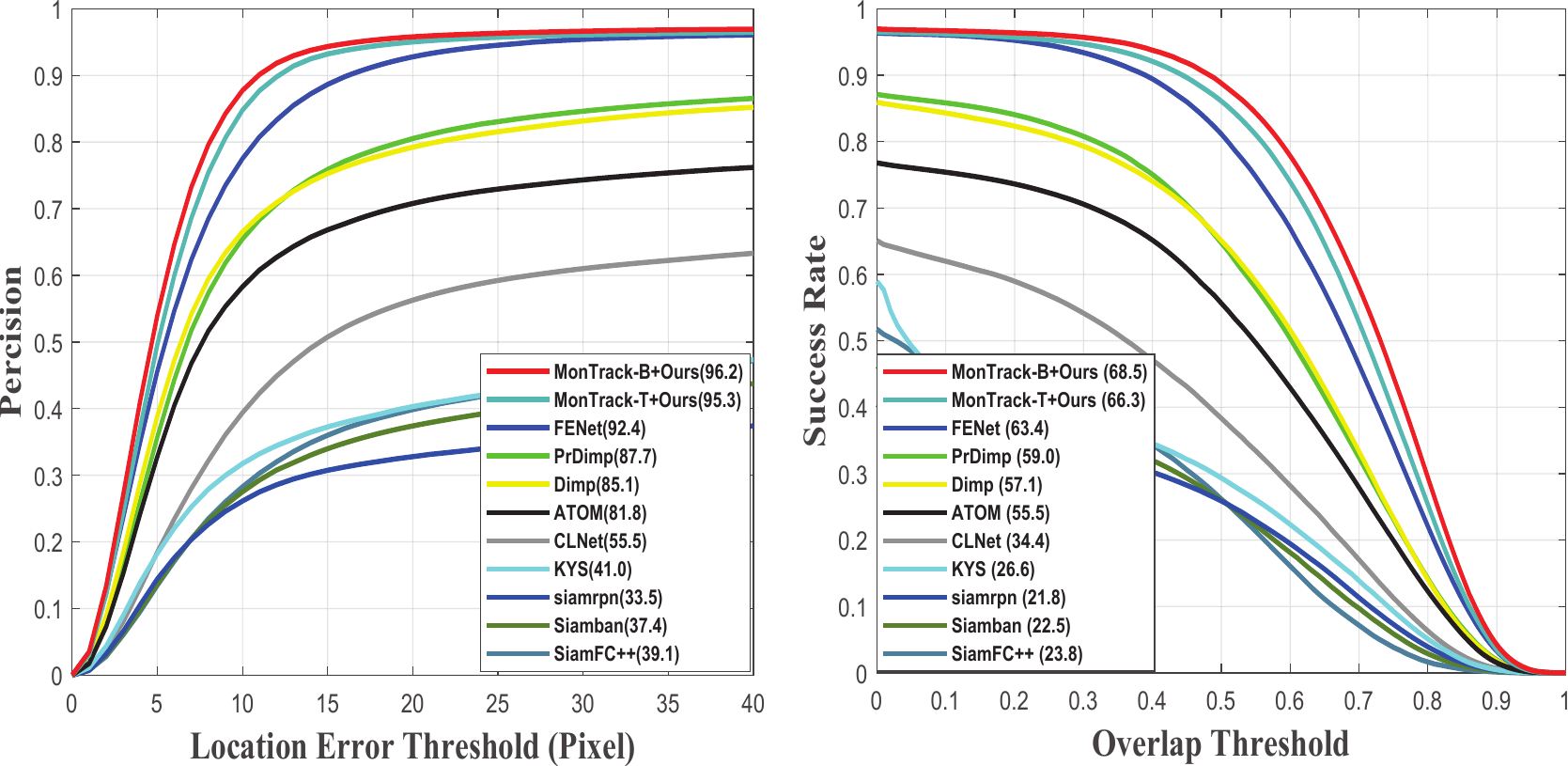}
  \caption{Precision and success plots of the FE108 dataset.} 
  \label{fig:Fe108_plot}
\end{figure}
\noindent \textbf{Datasets}. We employed two large-scale cross-modal RGB-event single object tracking datasets: FE108\cite{zhang2021object} and COESOT\cite{tang2022revisiting}. Both datasets were collected by \textit{DAVIS346} with a spatial resolution of 346 $\times$ 260, dynamic range of 120 dB, and minimum latency of 20 $\mu s$. FE108 consists of 108 RGB-event sequences collected indoors with a total length of 1.5 hours, which captures 21 different types of objects. The training split of FE108 consists of 140K RGB-Event pairs and 59K for testing. The ground-truth bounding boxes were annotated by a Vicon motion capture system. Moreover, the COESOT dataset consists of  578,721 RGB-Event pairs, which could be split into 827 and 527 sequences for training and testing, respectively. Those sequences are collected from both indoor and outdoor scenarios and cover a range of 90 classes and 17 attributes. The ground truth bounding boxes of the COESOT dataset were manually annotated. \textit{Note that we adopted the quantitative metrics suggested by each dataset to evaluate different methods}.

\begin{figure*}
  \centering
    \includegraphics[width =\textwidth, height= 0.26\textwidth]{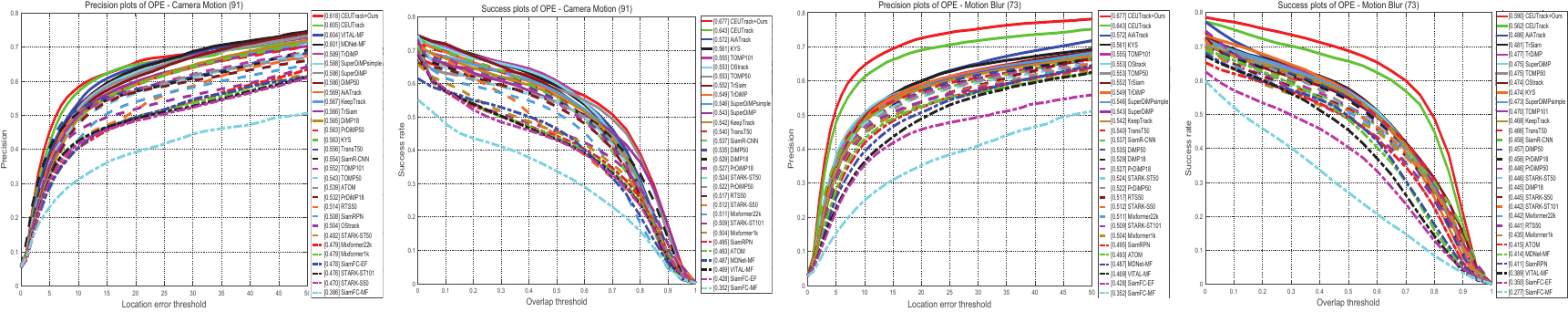}
  \caption{Visualization of the precision and success plots of sequences with different attributes in \textit{COESOT} dataset. The set of camera motion and motion blur contains 91 and 73 sequences, respectively. We also refer readers to the \textit{Supplementary Material} for comprehensive evaluations of all attributes \color{blue}{\faSearch~} Zoom in to see details.}
    \label{fig:attributes}
    \vspace{0.3cm}
\end{figure*}
\begin{figure*}
  \centering
    \includegraphics[ width=\textwidth, height= 0.6\textwidth]{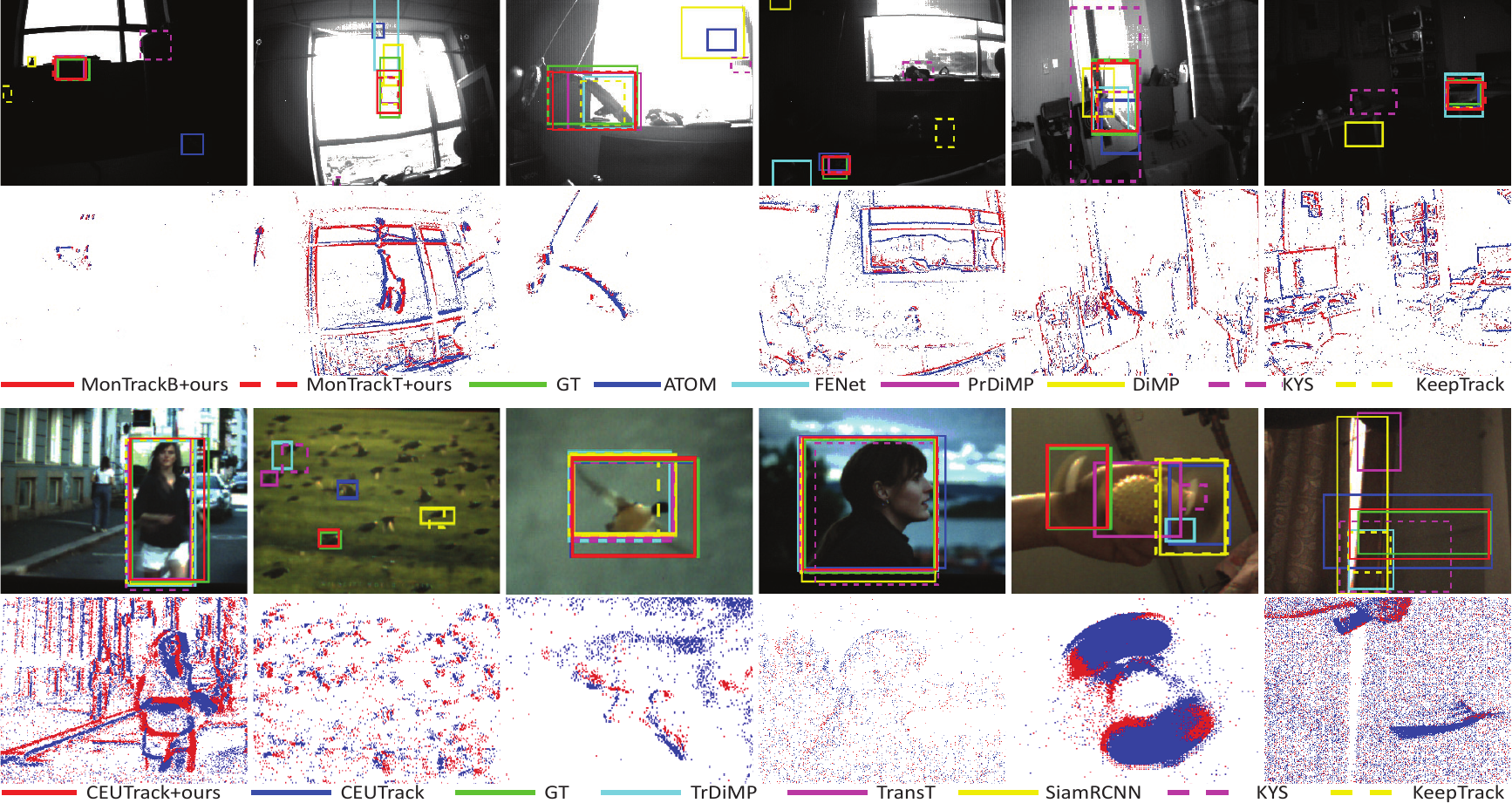}
  \caption{Visual comparisons of the tracking performance of different methods on the (\textbf{Upper}) FE108 and (\textbf{Bottom}) CEOSOT datasets. We also refer readers to the \textit{video demo} for more visual comparisons.}
    \label{fig:attributes}
\end{figure*}

\begin{table}
\setlength\tabcolsep{3pt}
  \caption{Quantitative comparison on the \textbf{COESOT} dataset in terms of four commonly-used metrics, i.e., success rate (SR), precision rate (PR), normalized precision rate (NPR), and breakOut capability score (BOC). For all metrics, the \textbf{larger}, the \textbf{better}. ``EVox" refers to the voxel representation of event data. } 
  \vspace{0.2cm}
  \label{table:COESOT}
  \centering
  \begin{small}
  
  \begin{tabular}{l | c |cccc}
    \toprule[1.2pt]
    Methods  & Modality & SR & PR & NPR & BOC \\
    \midrule
    \midrule
    RTS50~\cite{paul2022robust} & RGB + EI & 56.1 &62.6 &60.5 &16.88 \\    
    PrDiMP50~\cite{danelljan2020probabilistic} & RGB + EI & 57.9 &65.0 &64.0 &17.49\\
    KYS~\cite{bhat2020know} & RGB + EI & 58.6 &66.7 &65.7 &17.98\\
    DiMP50~\cite{bhat2019learning} & RGB + EI & 58.9  &67.1 & 65.9 & 18.07\\
    KeepTrack~\cite{mayer2021learning} & RGB + EI & 59.6 & 66.1 & 65.1 & 18.16\\
    TrSiam~\cite{wang2021transformer}              & RGB + EI &59.7 &66.3 &65.3 &18.15\\
    AiATrack~\cite{gao2022aiatrack} & RGB + EI & 59.0 &67.4 &65.6 &19.02\\
    OSTrack~\cite{ye2022joint} & RGB + EI & 59.0 &66.6 &65.4 &18.63\\
    ToMP101~\cite{mayer2022transforming} & RGB + EI  &59.9 &67.2&66.0&18.25 \\
    TrDiMP~\cite{wang2021transformer}  & RGB + EI & 60.1 &66.9 & 65.8 & 18.45\\
    TransT~\cite{chen2021transformer}  & RGB + EI & 60.5 &67.9 & 66.6 & 18.50\\
    SuperDiMP~\cite{danelljan2019pytracking} & RGB + EI & 60.2 &67.0 &66.0 &18.53\\
    SiamR-CNN~\cite{voigtlaender2020siam} & RGB + EI & 60.9  & 67.5& 66.3 & 19.08\\
    \midrule
    CEUTrack-B~\cite{tang2022revisiting} & RGB + EVox & 62.0&70.5 &69.0 &20.77\\
    CEUTrack-B+Ours & RGB + EVox &63.2 &71.9 &70.2 & 21.58\\
    \textbf{Improvement}     & -- -- & \textcolor{red}{+1.2} & \textcolor{red}{+1.4}& \textcolor{red}{+1.2} & \textcolor{red}{+0.81}\\
    \midrule
    CEUTrack-L~\cite{tang2022revisiting} & RGB + EVox & 62.8&71.4 &69.5 &20.97\\
    CEUTrack-L+Ours & RGB + EVox &65.0 &73.8 &71.9 & 22.40\\
    \textbf{Improvement}     & -- -- & \textcolor{red}{+2.2} & \textcolor{red}{+2.4}& \textcolor{red}{+2.4} & \textcolor{red}{+1.57}\\
    \bottomrule[1.2pt]
  \end{tabular}
  \end{small}
\end{table}


\subsection{Experimental Results}
\noindent\textbf{Results on FE108}. As listed in  Table~\ref{table:FE108}, after being augmented by the proposed techniques during training, both MonTrack-T and MonTrack-B substantially improve both RSR and PRP by more than $3\%$.  Moreover, the larger model ``MonTrack-B" yields a greater performance gain. We reason such an effect may be the consequence of promoting thoroughly cross-modal interaction  Besides, the superior performance of the proposed techniques is also demonstrated in the precision and success plots in  Fig.~\ref{fig:Fe108_plot}, which exceeds SOTA methods by a large extent, i.e., 5.1$\%$ in  RSR, 8.1$\%$ in OP$_{0.50}$, 12.1$\%$ in OP$_{0.75}$, and 3.8\% in RPR. Additionally, the higher performance of cross-modal  methods 
than that of only event-based methods and only RGB-based methods 
demonstrates the significance and necessity of using the information of both RGB and event data for object tracking.

\vspace{0.2cm}
\noindent\textbf{Results on COESOT}. As shown in Table~\ref{table:COESOT}, the original Tansformer-based cross-modal tracker, i.e., CEUTrack, improves the SR value of the previous SOAT SiamR-CNN by 1.1$\%$. After being augmented with our techniques, i.e., CEUTrack+Ours, the values of SR and PR are further improved by 1.2$\%$ and 1.4$\%$, respectively, and its NPR achieves higher than 70$\%$,
convincingly validating the effectiveness of the proposed techniques. In addition, we also provide the success and precision plots of different attributes in Fig.~\ref{fig:attributes}, where it can be seen that the proposed augmentations can yield general improvements instead of only strengthening certain circumstances. For example, 
the proposed augmentations achieve 3.4 $\%$ precision and 2.8 $\%$ success improvements under the blurring attribute. Especially,  CEUTrack+Ours maintains the best performance under the camera motion attribute, while the baseline CEUTrack drops to the 7$^{th}$. 

We also refer readers to the \textit{Supplementary Material} for the comparisons of the network size and inference time.

\begin{figure}
  \centering
    \includegraphics[width=0.48\textwidth]{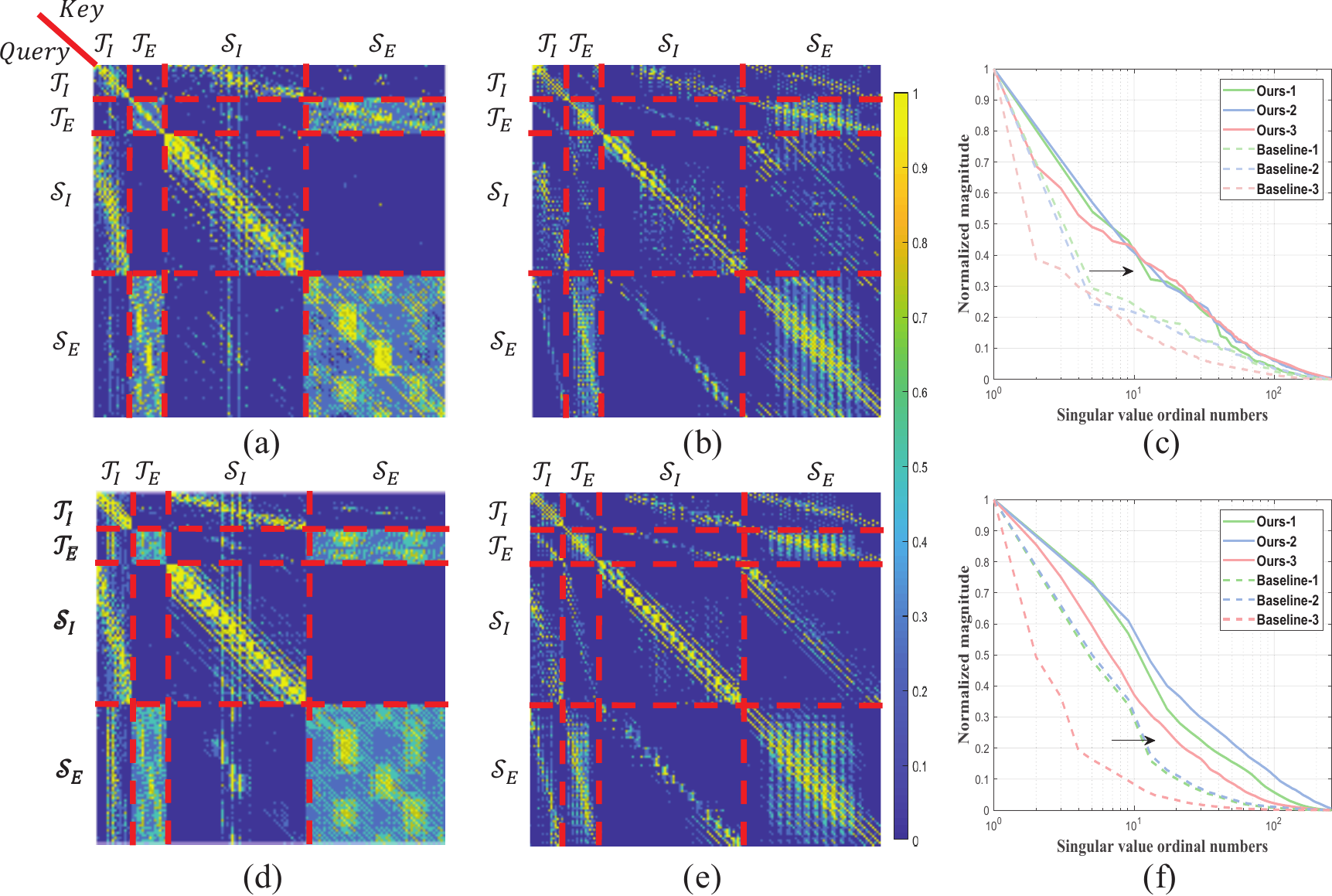}
  \caption{Visualization of the internal attention matrix, where we averaged six samples. (\textbf{a}) and (\textbf{d}) (resp. (\textbf{b}) and  (\textbf{e})) show the attention matrices of $5^{th}$ and $6^{th}$ attention layers of the baseline model CEUTrack (resp. the augmented model CEUTrack+Ours), respectively. 
  (\textbf{c}) (resp. (\textbf{f})) depicts the singular values of block matrices in (\textbf{a}) and (\textbf{b}) (resp. (\textbf{d}) and (\textbf{e}))  with ``*-1", ``*-2", and ``*-3" indicating the block matrices of $M_{\pazocal{T}_I,\pazocal{T}_E}$, $M_{\pazocal{T}_I,\pazocal{S}_E}$, and $M_{\pazocal{S}_I,\pazocal{S}_E}$, respectively. 
   We normalize both dimensions of curves in (\textbf{c}) and (\textbf{f}) into an identical range for better visualization.  Note that we arranged these matrices with the same manner as that in Fig.~\ref{fig:desirable}(\textcolor{red}{b}); hence, the summation of each row equals 1. See the supplementary material for more visualizations of the orthogonal property.}
    \label{fig:attn_matrix}
\end{figure}

\begin{figure}
  \centering
    \includegraphics[width=0.48\textwidth]{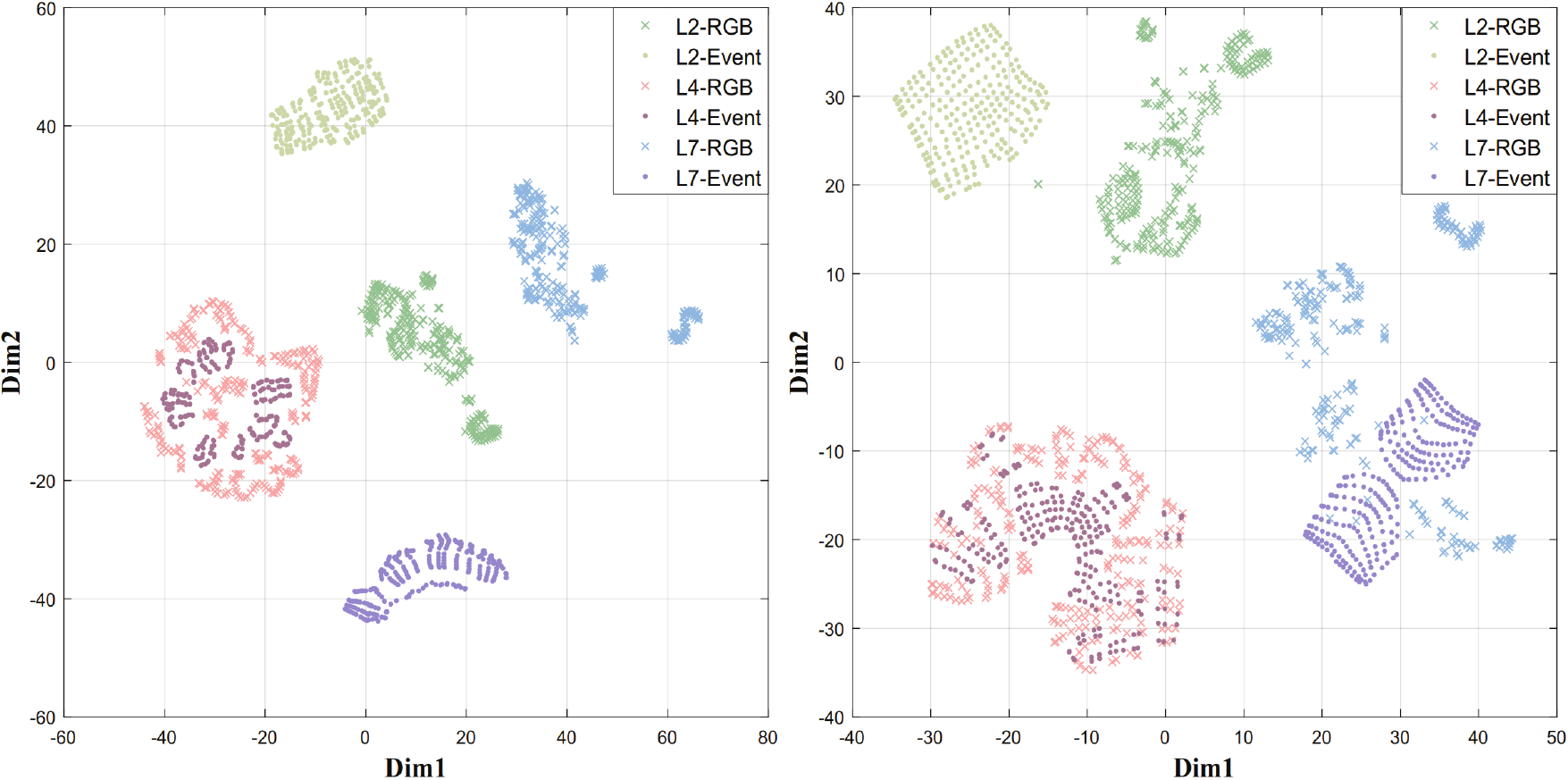}
  \caption{ t-SNE visualization~\cite{van2008visualizing} of the \textbf{query vector} in self-attention with the $2^{nd}$, $4^{th}$, and $7^{th}$ layers.  Visualization of the tokens from (Left) the baseline model CEUTrack and (Right) CEUTrack augmented with the proposed techniques during training.} 
    \label{fig:tsne}
\end{figure}

\subsection{Ablation Study} 
\label{sec:ablation}
\vspace{0.2cm}
\noindent\textbf{Visualizations}. Fig.~\ref{fig:attn_matrix} visualizes the internal attention matrix of CEUTrack. The values of each row of the  matrix are utilized to weight-sum tokens in that row and project to a corresponding token. Due to the absence of values in the blocks  $M_{\pazocal{S}_I,\pazocal{S}_E}$, $M_{\pazocal{S}_I,\pazocal{T}_E}$, $M_{\pazocal{T}_I,\pazocal{T}_E}$, $M_{\pazocal{T}_I,\pazocal{S}_E}$ in Figs.~\ref{fig:attn_matrix} (\textcolor{red}{a}) and (\textcolor{red}{d}), there is scarce information projected from the event domain to the RGB domain. 
The reason may be that the ViT was pre-trained on ImageNet composed of RGB data, 
making it preferable to process RGB data. 
When used as the backbone for constructing RGB-event object tracking, the pre-trained filters attempt to project event information onto RGB tokens to complete the labor-intensive tasks of information fusion and processing, instead of the inverse projection. After being augmented with our techniques during training, the cross-modal interaction is 
 noticeably enhanced, i.e., the matrix blocks, which are zeros in Figs.~\ref{fig:attn_matrix} (\textcolor{red}{a}) and (\textcolor{red}{d}), exhibit attention values, as demonstrated in Figs.~\ref{fig:attn_matrix} (\textcolor{red}{b}) and (\textcolor{red}{e}). 
 Besides, we also visualized the singular values of matrix blocks related to the cross-modal interaction in Figs.~\ref{fig:attn_matrix} (\textcolor{red}{c}) and (\textcolor{red}{f}), which substantially validates they have been pushed far away from a low-rank matrix after applying the proposed techniques. We refer readers to the \textit{Supplementary Material} for more results. 
Finally, Fig.~\ref{fig:tsne} shows the queries of the $2^{nd}$, $4^{th}$, and $7^{th}$ self-attention layers where it can be seen that the proposed augmentations narrow the distribution gaps between event and RGB tokens, especially for the $4^{th}$ layer.

\begin{figure}
  \centering
    \includegraphics[ width=0.5\textwidth]{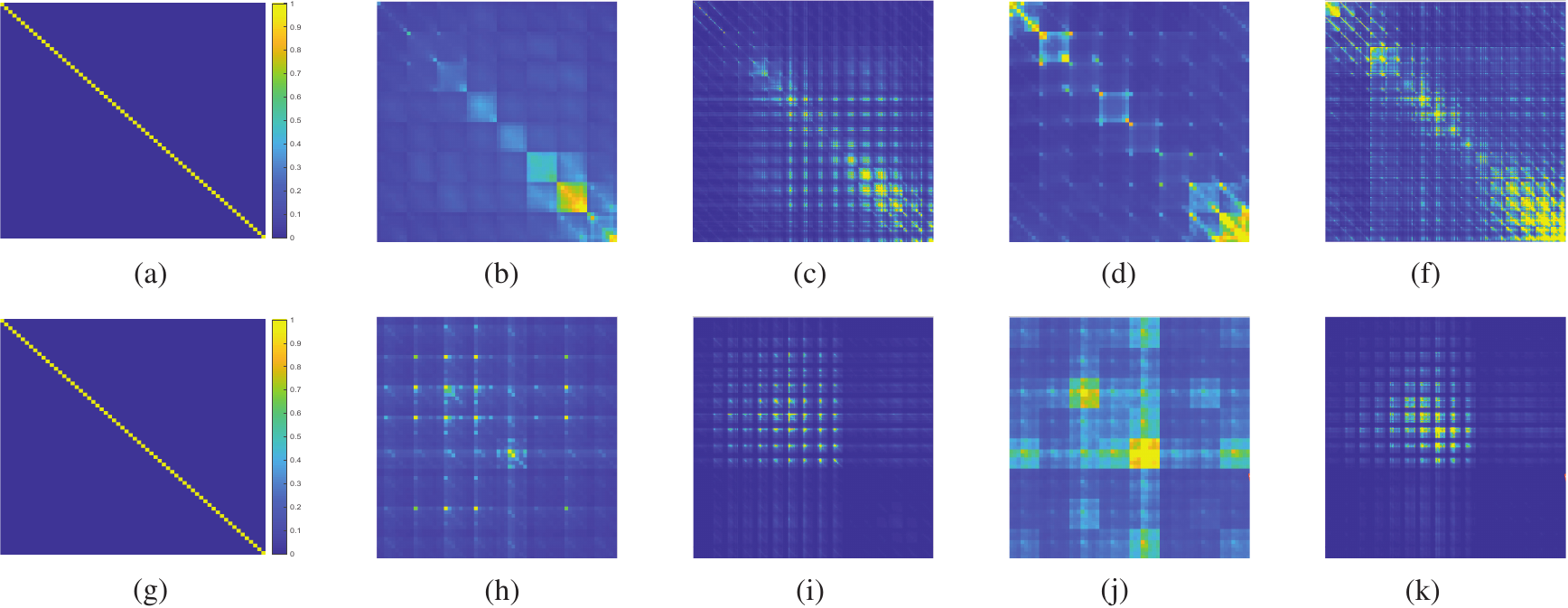}
  \caption{Visual comparisons of matrix orthogonality, where we visualize the matrix $A^\textsf{T}\cdot A$ of a given attention matrix $A \in \mathbb{R}^{n\times m}$ from CEUTracker on the COESOT dataset. (a) and (g) denote the fully-regularized orthogonal matrices, i.e., the optimization target. (b), (c), (d) and (f) (reps. (h), (i), (j) and (k)) indicate the matrices with (reps. without) the proposed augmentation scheme from $M_{T_E,S_I}$ of $5^{th}$, $M_{S_E,S_I}$ of $5^{th}$, $M_{T_E,S_I}$ of $6^{th}$, and $M_{S_E,S_I}$ of $6^{th}$ layer, respectively. }
    \label{fig:orthogonal1}
\end{figure}
\begin{figure}
  \centering
    \includegraphics[width=0.5\textwidth]{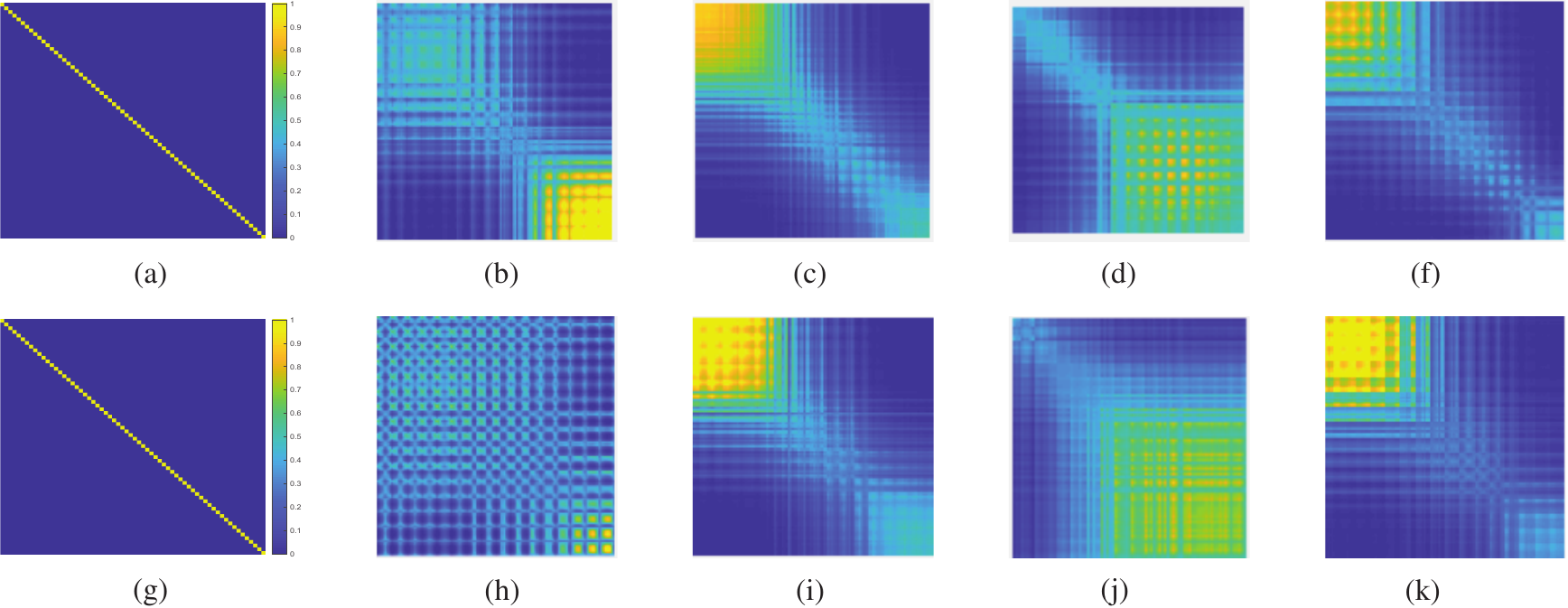}
  \caption{Visual comparisons of matrix orthogonality, where we visualize the matrix $A^\textsf{T}\cdot A$ of a given attention matrix $A \in \mathbb{R}^{n\times m}$ from MonTrack-B on the FE108 dataset. (a) and (g) denote the fully-regularized orthogonal matrices, i.e., the optimization target. (b), (c), (d) and (f) (reps. (h), (i), (j) and (k)) indicate the matrices with (reps. without) the proposed augmentation scheme from $M_{T,S}$ of $3^{th}$ layer $4^{th}$ head, $M_{T,S}$ of $3^{th}$ layer $5^{th}$ head, $M_{T,S}$ of $3^{th}$ layer $6^{th}$ head, and $M_{T,S}$ of $4^{th}$ layer $8^{th}$ head, respectively. }
  \vspace{-0.1cm}
    \label{fig:orthogonal2}
\end{figure}

\begin{table*}
\setlength\tabcolsep{3pt}
  \caption{Quantitative comparison of the model efficiency on the \textbf{COESOT} dataset. } 
  \label{table:COESOT}
  \centering
  \begin{small}
  \begin{tabular}{l | ccccccccccccccc}
    \toprule[1.2pt]
    Methods  &
    RTS50~&    
    PrDiMP50~ &
    KYS~&
    DiMP50~&
    KeepTrack~&
    TrSiam~&
    AiATrack~&
    OSTrack~& \\
    FPS  &
    30~&    
    30~ &
    20~&
    43~&
    18~&
    35~&
    38~&
    105~&\\
    \midrule
    Methods  &
    ToMP101~&
    TrDiMP~&
    TransT~&
    SuperDiMP~&
    SiamR-CNN~&
    CEUTrack~&
    CEUTrack+Ours \\
    FPS  &
    20~&
    26~&
    50~&
    -~&
    5~&
    75~&
    75\\
    \bottomrule[1.2pt]
  \end{tabular}
  \end{small}
\end{table*}

\noindent \textbf{Additional visualization of attention matrices}. To further directly visualize the variations of matrix orthogonality after applying the proposed augmentation, we visualized the matrix $M^\textsf{T}\times M$ in Figs.~\ref{fig:orthogonal1} and \ref{fig:orthogonal2}, convincingly demonstrating the effectiveness of our regularization for enhancing a matrix's row-/column-wise orthogonality.

\noindent\textbf{Masking} \textit{vs.} \textbf{High-rank}. We conducted throughout experiments to better understand the relationship and function of the proposed two augmentation techniques. From Table~\ref{table:Ablative_loss}, it can be seen that 
when the two techniques were simultaneously applied, the improvement is much more significant than that of only applying the masking scheme. The improvement is slight when only the high-rank regularization was applied.  These observations validate our claim that the two techniques are complementary.  

\noindent \textbf{Effect of the mask size}. We experimentally validated the effect of different mask sizes on performance. As shown in Table~\ref{table:MaskSize}, the benefits may be nullified under extremely large or tiny masks. The possible reason is that the network experiences the small masks as noise. While if the mask is too broad, the object may only appear in one modal, which may be detrimental to cross-modal learning.

\noindent \textbf{Model Efficiency}. We list the speed of different methods for comparison. Note that the proposed augmentation techniques were applied to baseline models only during the training phase. Consequently, no extra computational burden is imposed on the baseline models during testing. For the training phase, the CEUTrack operates at a rate of 30.6 samples per second. Moreover, it executes at 28.8 samples per second after plugging the presented regularization term.

\subsection{Discussion}
 In view of the impressive performance of the proposed plug-and-play training augmentations, it is worth further exploring their potential in other cross-modal scenarios, such as RGB-3D point clouds, or even vision-natural language. In addition, as demonstrated in Fig.~\ref{fig:attn_matrix}, the proposed orthogonal high-rank regularization indeed facilitates the interactions between cross-modal tokens, and thus, it would be promising to further develop task-specific regularization terms for other visual Transformers-based works.

\begin{table}
\setlength\tabcolsep{3pt}
  \caption{Results of the ablative study on the effect of the proposed two training augmentation techniques.} 
  \label{table:Ablative_loss}
  \centering
  \resizebox{.48\textwidth}{!}{
  \begin{tabular}{l | c c |cccc}
    \toprule[1.2pt]
    Baseline  & Masking & High-rank & RSR & OP$_{0.50}$ & OP$_{0.75}$ & RPR \\
    \midrule
    \midrule
    \multirow{4}{*}{MonTrack-B} &$\times$&$\times$&64.3 & 84.7 & 35.2 & 93.2 \\
    &$\checkmark$&$\times$ &67.6 &88.6 &41.8 &96.6\\
     &$\times$&$\checkmark$ &65.4 &85.9 &37.8 &94.6\\
     &$\checkmark$&$\checkmark$&68.5 & 89.4 & 45.4 & 96.2 \\
    \midrule
     Baseline & Masking & High-rank  & SR & PR & NPR & BOC \\ \hline
   \multirow{4}{*}{CEUTrack}  &$\times$&$\times$& 62.0 & 70.5 & 69.0 & 20.77 \\
    &$\checkmark$&$\times$ &62.4 & 71.0  &69.3 & 21.13 \\
     &$\times$&$\checkmark$ &61.7 & 70.2 & 68.6 &20.90\\
     &$\checkmark$&$\checkmark$ &63.2 & 71.9 & 70.2& 21.58\\
    \bottomrule[1.2pt]
  \end{tabular}
 }
\end{table}

\begin{table}
\setlength\tabcolsep{3pt}
  \caption{Results of the ablative study on the mask size.} 
  \label{table:MaskSize}
  \centering
  \begin{small}
  
  \begin{tabular}{l | c |cccc}
    \toprule[1.2pt]
    Methods  & Mask size &  RSR & OP$_{0.50}$ & OP$_{0.75}$ & RPR \\
    \midrule
    \midrule
    \multirow{4}{*}{MonTrack-T} &$1/2$ & 65.0 & 84.2 & 37.9 & 94.3\\
    &$1/4$ &64.8 & 84.6 & 34.0 & 95.8\\
    &$1/8$ &66.3 & 86.4 & 40.0 & 95.3 \\
    &$1/16$&64.2 &84.5 & 35.5 & 93.6\\
    \bottomrule[1.2pt]
  \end{tabular}
  \end{small} 
  \vspace{-0.1cm}
\end{table}

\section{Conclusion}
In this paper, we introduced plug-and-play training augmentations for Transformer-based RGB-event object tracking. Our augmentations consist of two complementary techniques--cross-modal mask modeling and orthogonal high-rank regularization with the same objective of enhancing the cross-modal interaction of a  ViT pre-trained only with RGB data. 
Our extensive experiments demonstrate the effectiveness of our training augmentations, as state-of-the-art methods achieve significant improvement in tracking performance after augmentation.

 While current Transformers can be scaled up to enormous sizes,  relying solely on final objectives to guide the model learning process may be insufficient. We hope our perspectives, findings and analysis 
 will inspire further research into the \textit{internal mechanisms} of Transformer-based cross-modal fusion tasks.

{\small
\bibliographystyle{ieee_fullname}
\bibliography{egbib}
}

\end{document}